\documentclass[nonacm, sigconf]{acmart}
\acmSubmissionID{1817}
\AtBeginDocument{%
  }
    
\def\eg{{\textit{e.g.}}}
\def\etal{{\textit{et al.}}}
\def\ie{{\textit{i.e.}}}
\usepackage[table]{xcolor}
\setcopyright{acmlicensed}
\copyrightyear{2025}
\acmYear{2025}
\acmDOI{XXXXXXX.XXXXXXX}
\acmConference[SIGGRAPH Asia]{SIGGRAPH Asia Conference Papers}{Dec 15-Dec 18, 2025}{Hong Kong, China}
\acmISBN{978-1-4503-XXXX-X/2018/06}

\makeatletter
\def\@fnsymbol#1{\ensuremath{%
  \ifcase#1\or \dagger\or \ddagger\or \mathsection\or \mathparagraph\or \|\or
  **\or \dagger\dagger\or \ddagger\ddagger \else\@ctrerr\fi}}
\makeatother
\begin{document}

\title{StyleSculptor: Zero-Shot Style-Controllable 3D Asset Generation with Texture-Geometry Dual Guidance}

\author{Zefan Qu}
\email{zefanqu2-c@my.cityu.edu.hk}
\orcid{0009-0006-5347-6426}
\affiliation{%
  \institution{City University of Hong Kong}
  \city{Hong Kong SAR}
  \country{China}
}

\author{Zhenwei Wang}
\authornote{Corresponding authors.}
\email{zhenwwang2-c@my.cityu.edu.hk}
\orcid{0000-0003-0215-660X}
\affiliation{%
  \institution{City University of Hong Kong}
  \city{Hong Kong SAR}
  \country{China}
}

\author{Haoyuan Wang}
\email{cs.why@my.cityu.edu.hk}
\orcid{0000-0003-4392-1435}
\affiliation{%
  \institution{City University of Hong Kong}
  \city{Hong Kong SAR}
  \country{China}
}

\author{Ke Xu}
\email{kkangwing@gmail.com}
\orcid{0000-0001-5855-3810}
\affiliation{%
  \institution{City University of Hong Kong}
  \city{Hong Kong SAR}
  \country{China}
}

\author{Gerhard Hancke}
\email{gp.hancke@cityu.edu.hk}
\orcid{0000-0002-2388-3542}
\affiliation{%
  \institution{City University of Hong Kong}
  \city{Hong Kong SAR}
  \country{China}
}

\author{Rynson W.H. Lau}
\authornotemark[1]
\email{Rynson.Lau@cityu.edu.hk}
\orcid{0000-0002-8957-8129}
\affiliation{%
  \institution{City University of Hong Kong}
  \city{Hong Kong SAR}
  \country{China}
}






\renewcommand{\shortauthors}{Qu et al.}

\newcommand{\reffig}[1]{\textcolor{black}{Fig.~\ref{fig:#1}}}
\newcommand{\refsec}[1]{\textcolor{black}{Sec.~\ref{sec:#1}}}
\newcommand{\reftab}[1]{\textcolor{black}{Tab.~\ref{tab:#1}}}
\newcommand{\refeq}[1]{\textcolor{black}{Eq.~\ref{eq:#1}}}

\newcommand{\todo}[1]{\textcolor{red}{[\textbf{TODO:} #1]}}
\newcommand{\wzw}[1]{\textcolor{black}{#1}}
\newcommand{\why}[1]{\textcolor{black}{#1}}
\newcommand{\qzf}[1]{\textcolor{black}{#1}}
\newcommand{\modify}[1]{\textcolor{black}{#1}}
\newcommand{\code}[1]{\textcolor{blue}{#1}}
\begin{abstract}
\wzw{Creating 3D assets that follow the texture and geometry style of existing ones is often desirable or even inevitable in practical applications like video gaming and virtual reality.}
  \wzw{While impressive progress has been made in generating 3D objects from text or images, creating style-controllable 3D assets remains a complex and challenging problem.}
  In this work, we propose StyleSculptor, a novel \wzw{training-free approach for generating style-guided 3D assets from a content image and one or more style images.}
  \wzw{Unlike previous works, StyleSculptor achieves style-guided 3D generation in a zero-shot manner, enabling fine-grained 3D style control that captures the texture, geometry, or both styles of user-provided style images}.
  At the core of StyleSculptor is a novel Style Disentangled Attention (SD-Attn) module,
  which establishes a dynamic interaction between \why{the input content image and style image for style-guided 3D asset generation via a \wzw{cross-3D attention mechanism}, enabling stable feature fusion and effective style-guided generation.}
  \wzw{To alleviate semantic content leakage, we also introduce a style-disentangled feature selection strategy within the SD-Attn module}, which leverages the variance of 3D feature patches to disentangle style- and content-significant channels, allowing selective feature injection within the attention framework. With SD-Attn, the network can dynamically compute texture-, geometry-, or both-guided features to steer the 3D generation process. Built upon this, we further propose the Style Guided Control (SGC) mechanism, which enables exclusive geometry- or texture-only stylization, as well as adjustable style intensity control.
  StyleSculptor does not require prior training and enables instant adaptation to any reference models while maintaining strict user-specified style consistency.
  Extensive experiments demonstrate that StyleSculptor outperforms existing baseline methods in producing high-fidelity 3D assets. Code and video are at \url{https://StyleSculptor.github.io}.
\end{abstract}


\begin{teaserfigure}
    \vspace{-2mm}
  \includegraphics[width=\textwidth, trim=0 182 20 5, clip]{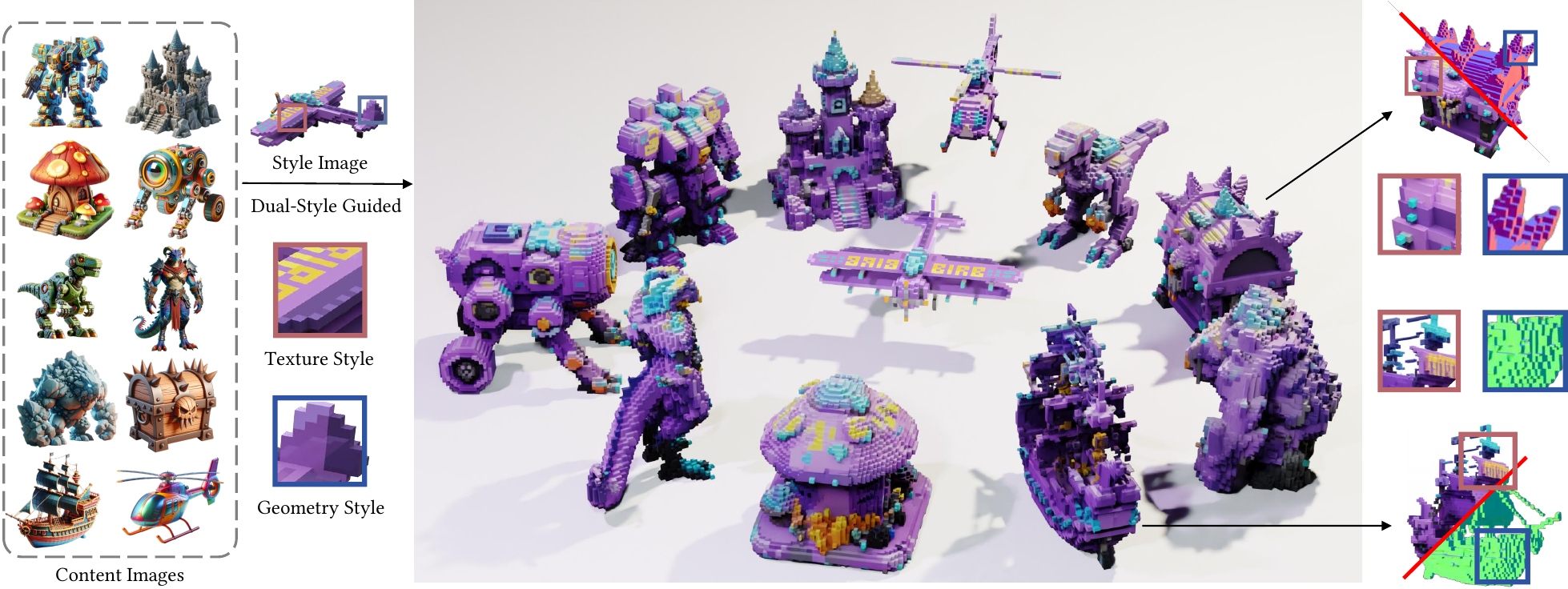}
  \vspace{-5mm}
  \caption{\qzf{We propose StyleSculptor, the first approach that can guide the generation of a novel 3D asset from a content image
  and reference \why{style images} in a zero-shot manner. 
 The results exhibit both geometric and textural style variations, while preventing unintended content leakage from the style image.}
  }
  \label{fig:teaser}
  \vspace{2mm}
\end{teaserfigure}


\maketitle

\section{Introduction} \label{intro}
\wzw{Maintaining style consistency is crucial for some practical 3D creative workflows, such as game design and virtual environments~\cite{wang2024themestation}. For instance, designers might need a voxelized warfare robot to match an existing voxelized plane for harmonious integration within a shared virtual world (Fig. ~\ref{fig:teaser}). 
We then consider if we could incorporate both texture and geometry style guidance into the automatic 3D generation process, thereby enhancing both efficiency and artistic controllability.
} 

Recent advances in AIGC and 3D representation~\cite{kerbl20233d, mildenhall2021nerf} have enabled large-scale pretrained models~\cite{chen2024comboverse,wu2024direct3d,lan2024ln3diff,li2024craftsman} that allow users to generate 3D assets from text or image input. \qzf{The advent of large-scale 3D generative models such as TRELLIS~\cite{xiang2024structured} and Hunyuan3D~\cite{zhao2025hunyuan3d} has significantly propelled advancements in the field of 3D generation.}
Although these methods have shown impressive performances, they lack control of the texture and geometry style during the generative process. 

\begin{figure}
    \vspace{-2mm}
  \includegraphics[width=0.5\textwidth, trim=0 265 400 0, clip]{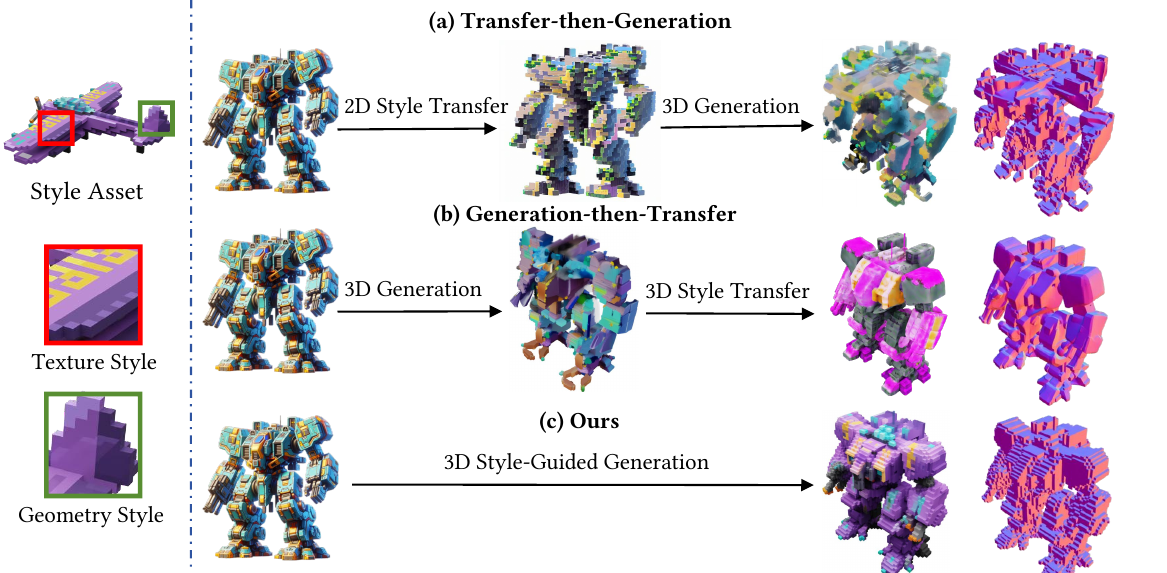}
  \vspace{-6mm}
  \caption{Comparison with the two-stage 
  \wzw{approaches}. 
  }
  \label{fig:twostage}
  \vspace{-3mm}
\end{figure}

\qzf{Given a content image and style images as input, the style-guided 3D generation task aims to produce \wzw{a 3D asset that is stylistically consistent with the style image while semantically aligned with the content image}. As shown in Fig.~\ref{fig:twostage}, two naive approaches can partially address this challenging task in a two-stage process.}
(1) Transfer-then-Generate:
Apply 2D style transfer to the input image. However, \why{these methods~\cite{ye2023ip,chung2024style,zhang2023inversion}
focus on 2D features only, often distorting 3D structure and geometry.} This can lead to \textbf{semantic inaccuracies}, hindering 3D reconstruction.
(2) Generate-then-Transfer: Performing 3D style transfer on the generated assets. Although these methods~\cite{liu2024stylegaussian,fujiwara2024style,xie2024styletex} can modify texture information, they often \textbf{do not adapt the geometric structure} of the asset to maintain 3D consistency, which limits applicability.


To address these limitations, in this work, we seek \why{a direct style fusion method} during 3D generation as a compelling alternative to conventional two-stage approaches, by leveraging the inherent ability of a pretrained 3D \qzf{rectified flow model~\cite{xiang2024structured}} to jointly model texture and geometry information. Our key observation reveals that when content and style features are properly integrated within the generative process, the network can produce style-consistent assets while preserving structural coherence. However, it is non-trivial to jointly inject both geometry and texture style guidance into the generation process in a zero-shot manner, given two critical challenges. \textbf{1) Semantic conflicts}. Simple feature fusion often fails when significant semantic disparities exist between content and style inputs, as the network struggles to interpret the coupled features. \textbf{2) Content leakage}. Uncontrolled transfer may lead to content leakage, where excessive style information corrupts the original asset's semantic integrity.

To overcome these challenges, we propose StyleSculptor, a novel zero-shot framework for texture and geometry style-guided 3D asset generation.
Given a content image and \why{style images}, StyleSculptor generates a 3D asset that preserves the semantic structure of the content while inheriting the texture and geometry style of \why{the style image}. 
Our solution performs implicit 3D-level style transfer by progressively fusing content and style features throughout the generation process. 
%

\wzw{To alleviate semantic conflicts between content and style, we propose a novel \textbf{Style-Disentangled Attention (SD-Attn)}  module, which employs \textit{1) Cross-3D attention,} which replaces the original self-attention with cross-attention between the intermediate features of two \why{conditions --- a content image and a style image}. This design maintains feature space stability during fusion, enabling successful style transfer even for semantically unrelated inputs. \textit{2) Style-disentangled feature selection strategy,} which leverages the variance of the 3D intermediate feature patches as additional guidance cues and selectively filters style-significant feature channels within the cross-3D attention framework. This strategy significantly} prevents unintended content distortion while allowing independent control over texture or geometry style transfer.

In addition to image-based style-\why{guided} 3D generation, StyleSculptor is also capable of performing style transfer on existing 3D assets.
By iteratively applying StyleSculptor with controlled activation of the SD-Attn module, we introduce a novel mechanism termed Style Guided Control (SGC). This mechanism enables fine-grained control over the style transfer process, allowing users to adjust the intensity of stylization, or to apply guidance exclusively to either geometry or texture. SGC significantly enhances the flexibility and controllability of the StyleSculptor network.


To evaluate the performance of our approach, we construct a dataset comprising content images and 3D style assets across various objects and styles, and test numerous SOTA methods on this dataset. Our contributions can be summarized as follows.
\begin{itemize}
\item[$\bullet$] We propose StyleSculptor, a fine-grained training-free style-guided 3D asset generation \why{method} that enables simultaneous control over both texture and geometry. 
\item[$\bullet$] We propose the Style Disentangled Attention (SD-Attn) module, which integrates cross-attention for content-style fusion and leverages channel-wise feature variance to decouple style and content information, enabling precise style transfer without content leakage.
\item[$\bullet$] We propose Style Guidance Control (SGC), a flexible mechanism that enables fine-grained control over style intensity and disentangled geometry- or texture-only stylization, significantly enhancing the controllability of the framework.
\end{itemize}

\section{Related Work}
\subsection{3D Generative Models}
In recent years, considerable effort~\cite{luo2021diffusion, nichol2022point, hui2022neural,muller2023diffrf, chen2023single, shue20233d, poole2022dreamfusion,wang2024themestation} has been directed to extend the advantages of 2D diffusion models~\cite{ho2020denoising, rombach2022high} to 3D generation tasks. 
\modify{In pursuit of improved quality and efficiency, recent works~\cite{ren2024xcube, zhang20233dshape2vecset, zhang2024clay, chen2019net, hertz2022spaghetti} have increasingly focused on generation within a more compact latent space.} Several methods have concentrated primarily on shape modeling, often requiring a separate texturing step to complete the creation of 3D assets. 
\modify{The advent of large-scale 3D generative models~\cite{zhao2025hunyuan3d,xiang2024structured,wu2024direct3d, chen20253dtopia,hui2024make, li2025triposg} has significantly propelled advancements in the field of 3D generation.}
Through optimization on extensive datasets, they produce highly expressive 3D latent features, which can fully express the texture and geometric layout of a 3D asset.
These generation pipelines can only generate the corresponding asset based on the input, without the ability to effectively guide the generation process.

\subsection{Controllable 3D Generation}
Several approaches have emerged that aim to provide image-based guidance for the 3D generation process. For example, some approaches~\cite{xu2023dream3d, huang2024dreamcontrol,dong2024coin3d,wang2024phidias,gao2025charactershot} focus on incorporating shape priors to influence the geometric properties of the generated content. Meanwhile, some methods~\cite{richardson2023texture,yeh2024texturedreamer, xie2024styletex, song2024style3d} enable the unification of the style of the generated object by leveraging the reference images. These style-guided 3D generation techniques typically require users to provide 3D information, such as a mesh, while keeping the geometric structure fixed throughout the generation process to preserve spatial consistency. This constraint limits their ability to control only the texture style of the generated asset, leaving the geometry style, which is also crucial for 3D assets unaltered and uncontrollable.

 In this work, we adopt TRELLIS~\cite{xiang2024structured} as our baseline model, leveraging its expressive 3D latent features, which comprehensively encode both the geometry and texture of the generated objects. Building on these features, we can achieve both texture and geometry style-guided 3D asset generation in a zero-shot manner.

\subsection{Image-based Style Transfer}
Image-based style transfer involves extracting the visual style from source images and applying them to a target content, which can be a 2D image or a 3D model, while preserving the content information. 

\textbf{2D Style Transfer}. 
With the advent of diffusion models, many recent methods have leveraged their powerful generative capabilities to achieve more compelling and robust style transfer results. Zhang \etal~\shortcite{zhang2023inversion} propose an efficient textual inversion method for feature learning and content preservation. 
Chung \etal~\shortcite{chung2024style} and Ye \etal~\shortcite{ye2023ip} achieve style transfer by manipulating  the image diffusion process.
These methods are limited to 2D style modifications, directly applying them to alter input images for 3D tasks may disrupt the latent spatial properties of the images, ultimately causing the 3D generation process to fail.

\textbf{3D Style Transfer}. Style transfer task is much more complex in 3D than 2D because it includes both texture and geometry style. Fan \etal~\shortcite{fan2022unified} and Liu \etal~\shortcite{liu2023stylerf} propose the network that can transfer the color tone of the style images to the NeRF scenes. Jain \etal~\shortcite{jain2024stylesplat} propose a feature matching loss to adjust the color of the selected objects. Liu \etal~\shortcite{liu2024stylegaussian} designs an efficient feature rendering strategy to embed high-dimensional features into 3D Gaussians. Richardson \etal~\shortcite{richardson2023texture} and Xie \etal~\shortcite{xie2024styletex} generate textures for an existing 3D shape with the 2D generative model.
To avoid disruption of spatial consistency in the scene during the transfer process, these methods preserve the geometry of the 3D scene while only modifying its texture. Unlike 2D styles, many geometry-aware styles in reference images, such as pixel art or grid-like patterns—cannot be adequately represented by simply altering the color distribution. In this paper, we consider both the texture and geometry styles during the style-guided 3D generation process to obtain a more desirable result.

\begin{figure*}
\vspace{-3mm}
  \includegraphics[width=\textwidth, trim=0 120 0 95, clip]{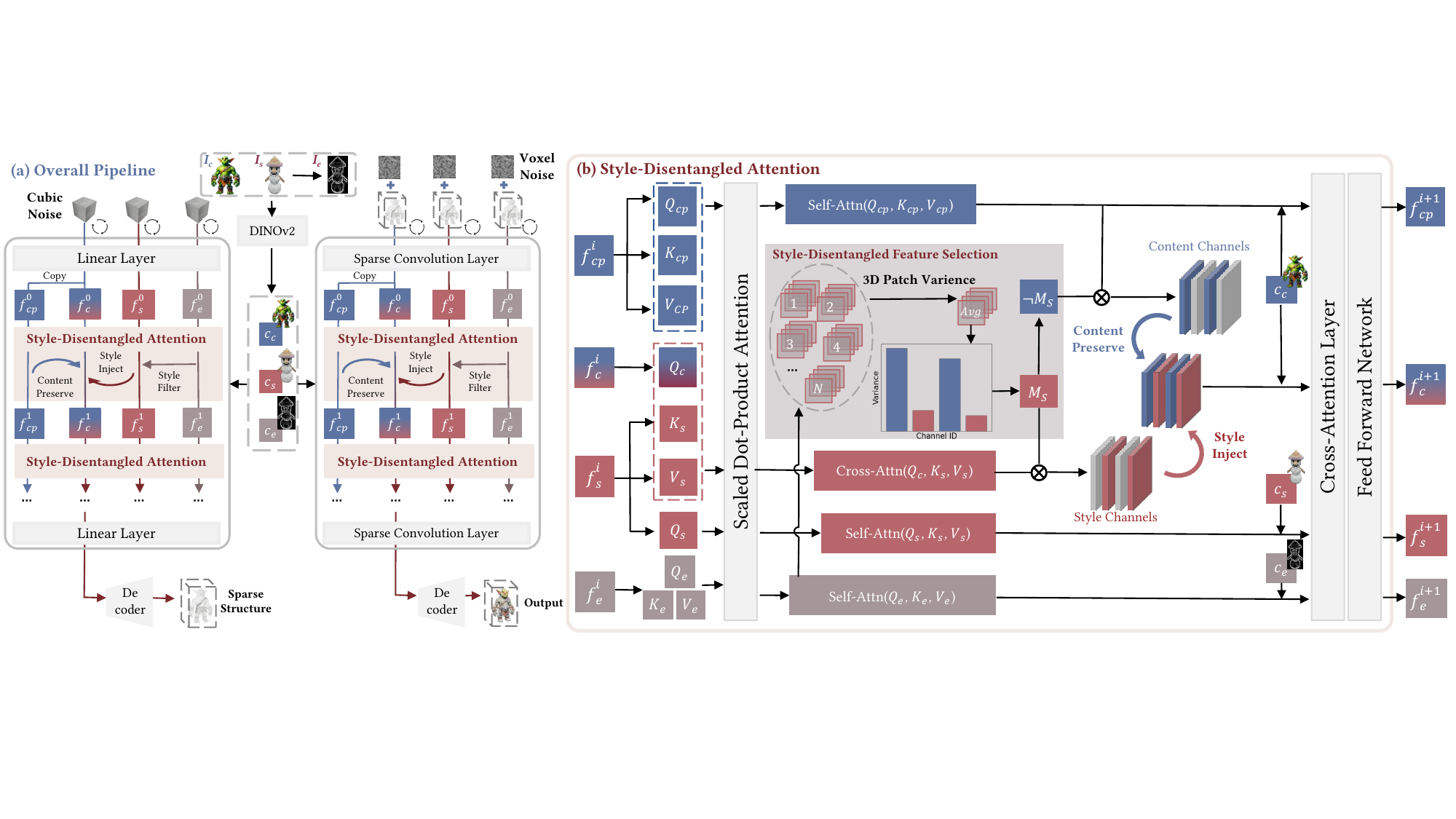}
  \vspace{-6mm}
  \caption{(a) Overall pipeline. Given a content image $I_c$ and one/more style images $I_s$, StyleSculptor generates a 3D asset that is stylistically consistent with the style image and semantically aligned with the content image in a zero-shot manner.
  (b) Style-Disentangled Attention (SD-Attn) module.
  It consists of a Cross-3D Attention mechanism and a Style-Disentangled Feature Selection (SDFS) strategy to achieve style-guided generation in a content-preserving yet style-consistent manner.
  }
  \label{fig:pipeline}
  \vspace{-2mm}
\end{figure*}

\section{Approach}
In this section, we present the architecture and mechanism of StyleSculptor that enables texture and geometry style-guided 3D generation. We start with an introduction of the preliminaries (Sec.~\ref{preliminary}), followed by the overview of our pipeline (Sec.~\ref{pipeline}). Then, we illustrate the proposed Style-Disentangled Attention (SD-Attn) module (Sec.~\ref{sa-att}), which facilitates style transfer while preserving content semantics. Finally, a stylization controllable mechanism named Style Guided Control (SGC) is introduced (Sec.~\ref{control}) to control the guidance during the generation process.

\subsection{Preliminaries}
\label{preliminary}
\subsubsection{TRELLIS Pipeline}

TRELLIS~\cite{xiang2024structured} is a robust foundation model for 3D asset generation that leverages user-provided text or images to create 3D models in various representations (\eg, NeRF, 3DGS, Mesh).
It employs a two-stage generation process: \why{first, it generates the asset's sparse structure, and then produces detailed latent features conditioned on this sparse structure.} In both stages, TRELLIS utilizes the \why{Rectified Flow Model to fit the feature distribution.}

In the first stage, the DINOv2 model~\cite{oquab2023dinov2} is used to extract the feature $ c $ from the input images, which serves as the condition for \why{both stages} of the network. A 3D noise feature grid $ S_t $ is then initialized and input into the Flow Transformer $ V_{FT} $ for noise prediction, as follows:
\begin{equation}
S_{t-\Delta t} = S_t - V_{FT}(c, S_t, t) \, \Delta t \label{step1},
\end{equation}
where \( t \) and $\Delta t$ represent the time variable and time step length. After obtaining $ S_0 $, the decoder processes it to generate the voxels of the sparse structure of the asset, denoted as $ \{p_i\}_{i=1}^L $, where $ L $ is the activate number of voxels. For each voxel, a noise latent $ \{Z_i\}_{i=1}^L $ is initialized and input into the Sparse Flow Transformer ($ V_{SFT} $) for the second stage of denoising, as follows:
\begin{equation}
Z_{t-\Delta t} = Z_t - V_{SFT}(c, p, Z, t) \, \Delta t \label{step2}.
\end{equation}
Then, by processing this latent 3D representation through different decoders, the network can generate various types of 3D asset model. 


In this work, we adopt TRELLIS~\cite{xiang2024structured} as the backbone of StyleSculptor, where the 3D latent features from both stages, denoted collectively as $ f $, serve as the key representation for 3D generation and style injection, due to their strong content- and style-aware modeling capabilities.
The attention mechanisms in $ V_{FT} $ and $ V_{SFT} $ are further adapted to enable \textbf{texture and geometry-guided style transfer} in a zero-shot manner.
\modify{Specifically, we replace all the self-attention layers in the Flow Transformer (first stage) and the Sparse Flow Transformer (second stage) of TRELLIS with our SD-Attn module. The cross-attention and FFN layers in each transformer block are kept intact and only the self-attention computation is modified. For different inputs, the QKV features are derived from the original QKV projection matrices in TRELLIS.}

\subsection{Overview} \label{pipeline}

The overall pipeline of StyleSculptor is depicted in Fig.~\ref{fig:pipeline}(a). \wzw{Given} a content image $I_c$ and one or more style images $I_s$, we aim to generate a 3D asset consistent with the style images in texture, geometry, or both, \why{according to the user's preference.}
To achieve style-guided generation in a content-preserving yet style-consistent manner, we propose the Style-Disentangled Attention (SD-Attn) module with cross-3D attention and a style-disentangled feature selection strategy~(Sec.\ref{sa-att}). \why{The cross-3D attention operates on intermediate features of two potential 3D assets: a content 3D asset (implicitly derived from the content image) and a style 3D asset (implicitly derived from the style image)}. The style-disentangled feature selection strategy aims to filter out semantic features and utilize only the style-significant features to guide the 3D generation process. 
This strategy employs additional edge maps $I_e$ of each style image for 3D feature variance-based feature channel filter.
As shown in the Fig.~\ref{fig:pipeline}(a), StyleSculptor generates a style-guided 3D asset in two stages. \why{First, the content features $f_c$, style features $f_s$, and edge map-derived features $f_e$ are processed by the proposed SD-Attn modules. Within these modules, $f_e$ aids in filtering style features from $f_s$, and appropriate style information is then injected into the content generation pipeline (operating on $f_c$).}
After completing all denoising steps in the first stage, the content branch outputs a fully \wzw{style-guided} feature, which is decoded into a sparse structure. This structure is then re-noised and fed into the second stage, where the denoising process repeats to refine the output. \why{Both stages use the same initial noise} to ensure feature distribution consistency. Notably, no backbone parameters are modified during this process. Instead, zero-shot style guidance is achieved, making our method lightweight and flexible.



\vspace{-3mm}
\subsection{Style-Disentangled Attention Module} \label{sa-att}


\wzw{To generate a 3D asset that follows the semantics of the content image $I_c$ and styles of the style images $I_s$, a naive solution is combining} the feature of content $f_c$ and style $f_s$ to guide the image-to-3D generation process. However, this may introduce two key problems: 1) the mechanism becomes less effective~\cite{alaluf2023crossimage} when content and style differ significantly in semantics; 2) this may cause unintended content leakage from the style image $ I_s $, leading to undesirable outputs. To solve these two problems, we introduce cross-3D attention with a style-disentangled feature selection strategy.

\wzw{\textbf{Cross-3D Attention.}} To address the first issue, \wzw{we extend the cross-image attention techniques~\cite{chung2024style, alaluf2023crossimage} to cross-3D attention in our style-guided 3D generation pipeline.} 
\wzw{\why{Using the backbone model, \ie, TRELLIS, we observe that the intermediate features the network learned} during the image-to-3D generation process implicitly encoded the style features of the conditioned image. 
In style-guided 3D generation, replacing the self-attention layers with cross-3D attention layers \why{between the content and the style's intermediate features} can effectively guide the model's output and progressively fuse the content and style features.
\qzf{We perform two generation processes using both $ I_c $ and $ I_s $, and obtain their corresponding intermediate latent features at the same $t$, denoted as $ f_{c} $ and $ f_s $, respectively.}
Now, with both $f_c$ and $f_s$, we could finally obtain a target style-guided 3D asset by replacing the self-attention layers during the denoising process of $f_c$ with the cross-3D attention layers between $f_c$ and $f_s$, as follows:}
\begin{equation}
\text{Cross-Attention}(Q_c, K_s, V_s) = \text{softmax}\left(\frac{Q_c \cdot K_s^T}{\sqrt{d_k}}\right)V_s, \label{crossatt}
\end{equation}
where $Q_c$ is the query from $f_c$, $K_s$ and $V_s$ are the key and value from $f_s$. After this operation, $f_c$ still keeps the feature distribution uncorrupted during the fusion process. 
%

\wzw{\textbf{Style-Disentangled Feature Selection.}} To further alleviate the content leakage problem, our goal is to decompose $ I_s $ into semantic and stylistic components, retaining only the latter for fusion with the content features. 
Based on this idea, we propose a Style-Disentangled Feature Selection (SDFS) strategy, guided by two insights. 
\textit{1) 3D features channels can be divided into content- and style-significant channels.} With this insight, \why{we can then compute the cross attention only on those style-significant channels, preserving the information of the content image.}
\textit{2) Style information exhibits better global consistency than content.} The style information is consistent \why{across locations in the 3D asset, while} content information is relatively more variable. This allows us to use statistical metrics to separate content-significant and style-significant channels. 
\wzw{Experiments validate these two insights can be found in Sec.~\ref{sec:experiment}.}

Based on the two insights discussed above, we employ 3D feature variance among all the patches in each channel (3D-Var) as the statistical measure to classify channels into content- and style-dominant categories. Since style information tends to be globally consistent, the variance of feature values across patches within a channel is typically small for style-significant channels, \why{while being larger for content-significant ones.}
\wzw{However,} when the \why{style image} contains complex local color or geometric details, the effectiveness of channel classification based on 3D-Var may be unstable. To address this issue, we employ \why{edge maps of the style images as an additional condition to guide the generation process, using the extracted feature $f_e$ as the style channel filter}, which is less affected by intricate local patterns and contains the general semantic information compared to $ f_s $.
Specifically, in the self-attention operation, the feature $ f_{e} $ is processed as:
\begin{equation}
    f'_{e}{}^{(N \times C)} = \text{Self-Attention}(Q_{e}, K_{e}, V_{e}),
\end{equation}
where $ N $ and $ C $ denotes the number of patches and feature channels respectively. Then the variance $\text{3D-Var}^{(C)}$ in each channel can be calculated as:
\begin{equation}
    \text{3D-Var}^{(C)} = \frac{1}{N} \sum_{n=1}^{N} \left( f'_{e}{}^{(n,C)} - \mu \right)^2, \quad \text{where } \mu = \frac{1}{N} \sum_{n=1}^{N} f'_{e}{}^{(n,C)}.
\end{equation}


\begin{figure}
\vspace{-3mm}

  \includegraphics[width=0.5\textwidth, trim=0 350 475 0, clip]{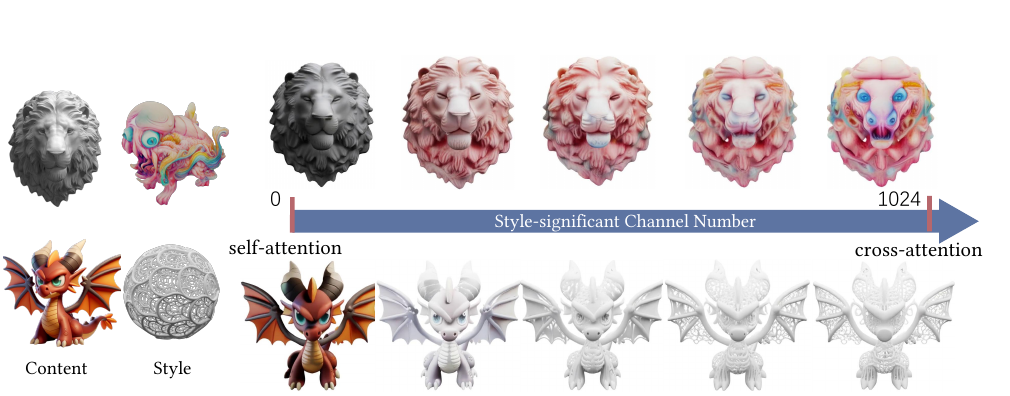}
  \vspace{-6mm}
  \caption{The \wzw{intensity of guidance} affected by the style-aware channel numbers.
  }
  \label{fig:channel-change}
   \vspace{-3mm}
\end{figure}

After computing the patch-wise variance for each channel, we sort the channels in order of variance and select the top-$ K $ channels with the smallest variances to construct a binary style-aware channel mask $ \mathcal{M}_s^{(C)} \in \{0,1\}^{(C)} $. Specifically, the selected channels are marked as 1, indicating strong style relevance, while the remaining are set to 0, corresponding to content-dominant channels.
Then we perform cross-attention between $ f_{c} $ and $ f_s $, and retain only the features in the style-significant channels. 
Additionally, we introduce a \textbf{Content Preserve} mechanism. Before each denoising step, we create a copy of $f_{c}$, denoted as $f_{{cp}}$, which undergoes the same SD-Attn forward pass with the normal self-attention calculation. Not having cross-attention computed makes $f_{cp}$ have a complete distribution of content information compared to fused feature $f_c$.
As a result, the remaining content-significant channels in SD-Attn module are complemented using self-attention features $ f_{cp} $. The overall computation is formulated as:
\begin{equation}
\begin{split}
    f'_{c} 
    &= \text{Cross-Attention}(Q_{c}, K_s, V_s) \otimes \mathcal{M}_s^{(C)} \\
    &\quad + \text{Self-Attention}(Q_{{cp}}, K_{{cp}}, V_{{cp}}) \otimes (1 - \mathcal{M}_s^{(C)}),
\end{split}
\end{equation}
where $ \otimes $ denotes channel-wise multiplication with broadcasting. The \wzw{cross-3D attention} mechanism and style-disentangled \wzw{feature selection} strategy in \wzw{our SD-Attn} module ensure that only style-relevant information is injected into the content branch at each step, preventing potential disruptions to the feature distribution and avoiding semantic content leakage during the stylization process. 

\subsection{Style Guidance Control Mechanism} \label{control}

Enabling controllable style-guided generation is of great importance for user customization. To this end, StyleSculptor introduces a flexible solution named Style Guidance Control (SGC) that allows users to control the generation process in three aspects: 
(1) the intensity of style guidance, 
(2) geometry-only \wzw{style guidance}, and 
(3) texture-only \wzw{style guidance}. 
This controllable generation is achieved simply by adjusting the hyperparameter $ K $, which determines the number of style-aware channels.


\begin{figure*}
  \includegraphics[width=0.95\textwidth, trim=0 165 40 10, clip]{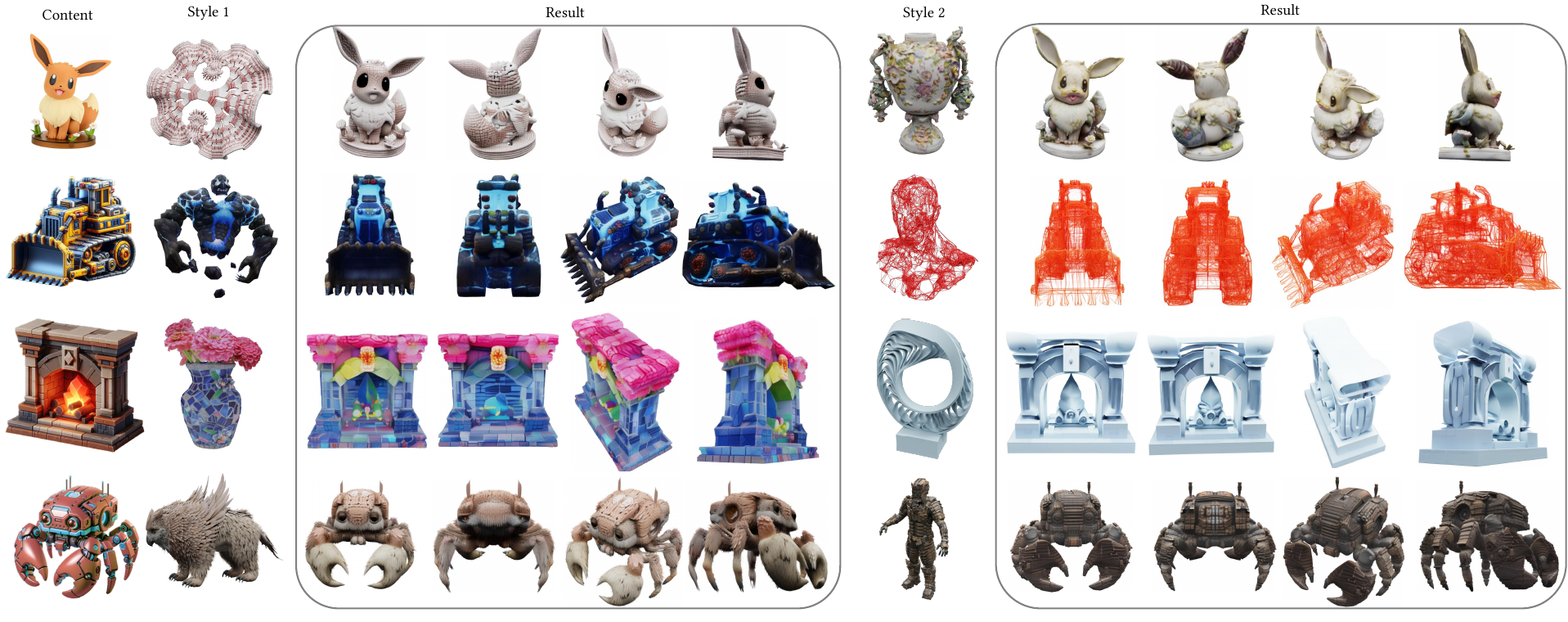}
  \vspace{-4mm}
  \caption{Qualitative results of Dual-style guided StyleSculptor. We present two style-guided generation results for each content.
  }
  \label{fig:qualresult}
  \vspace{-2mm}
\end{figure*}

\wzw{\textbf{Intensity
of Style Guidance.}} As shown in Fig.~\ref{fig:channel-change}, varying the value of $ K $ leads to different \wzw{intensities of style guidance}. When $ K = 0 $, the SD-Attn degenerates into a basic self-attention layer, and the generated output closely resembles the content image. In contrast, when $ K = 1024 $ (the maximum number of channels), \wzw{SD-Attn} behaves as a full cross-image attention module, resulting in complete style content leakage into the generated 3D asset.
Notably, as $ K $ increases, the degree of \wzw{style guidance} becomes visibly stronger. Even when $ K $ approaches the channel limit, StyleSculptor still preserves a significant amount of content semantics while fully stylizing the remaining regions.


\wzw{\textbf{Geometry-only and Texture-only Style Guidance.}} As $ K $ increases from 0, the generated asset first adopts texture information from the style image, followed by geometric styles. This observation provides a practical solution to achieve texture-only and geometry-only style guidance. Please see the detailed design of our single-style guided 3D generation in the supplementary material.




\section{Experiments} \label{sec:experiment}
\begin{figure*}
  \includegraphics[width=\textwidth, trim=0 370 40 0, clip]{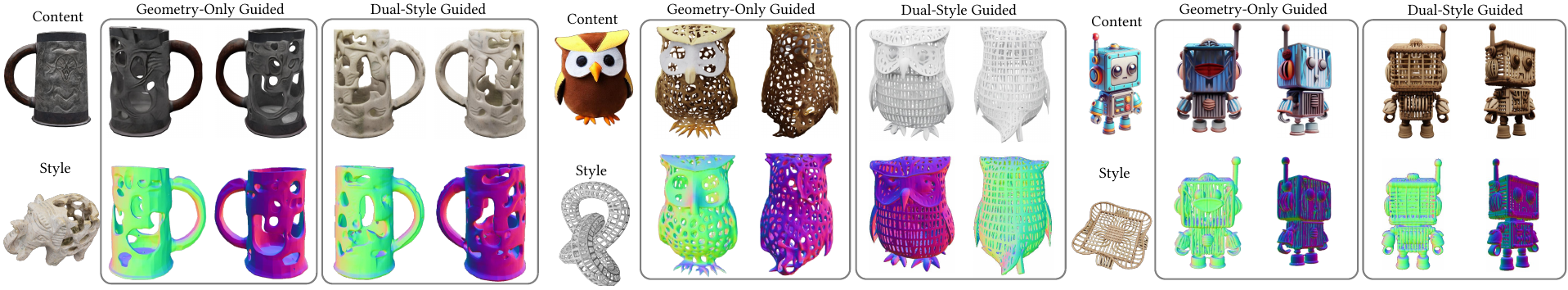}
  \caption{\modify{Results of geometry-only style-guided 3D generation.}
  }
  \label{fig:georesult}
\end{figure*}

\begin{figure*}
  \includegraphics[width=\textwidth, trim=0 370 28 0, clip]{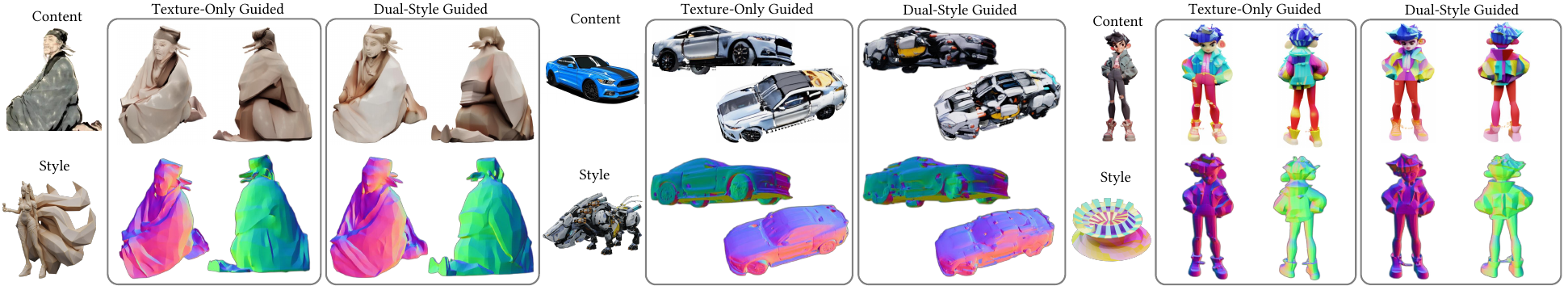}
  \caption{\modify{Results of texture-only style-guided 3D generation.}
  }
  \label{fig:textresult}
\end{figure*}

\wzw{We show the style-guided 3D generation results produced by StyleSculptor under three different settings, including geometry-only (\reffig{georesult}), texture-only (\reffig{textresult}), and texture-geometry dual guidance (\reffig{qualresult}). As can be seen, our approach can generate style-consistent 3D assets \why{given} various styles,} even when there is a significant semantic gap between the style and content image.
\wzw{For the rest of this section, we first conduct experiments and a user study to compare our results with those generated by existing SOTA methods. We then perform ablation studies to validate the effectiveness of each proposed component. We provide the implementation details in the supplementary material.}
\modify{Note that in this paper, we render the RGB views of the generated 3D assets from their Gaussian Splatting representation, and the normal maps from their Mesh representation.}

\subsection{Comparisons with State-of-the-Art Methods}

\subsubsection{Benchmark}
We collect 50 3D assets with distinct geometric and texture styles from ObjaverseXL~\cite{deitke2023objaverse} and Sketchfab\footnote{https://sketchfab.com. All assets attribution can be found in the Supplemental.} under creative commons licenses, \wzw{and render the assets} as style inputs. These assets include 25 objects, 13 characters, and 12 creatures, \wzw{covering various texture styles and exhibiting} prominent geometric characteristics such as lattice structures or pixel-art patterns. \wzw{For content inputs}, we select 30 public images with explicit 3D structure from the StyleBench~\cite{gao2024styleshot} and  TRELLIS~\cite{xiang2024structured}.

\subsubsection{Methods Compared.} 
\begin{figure*}
  \includegraphics[width=\textwidth, trim=0 220 30 0, clip]{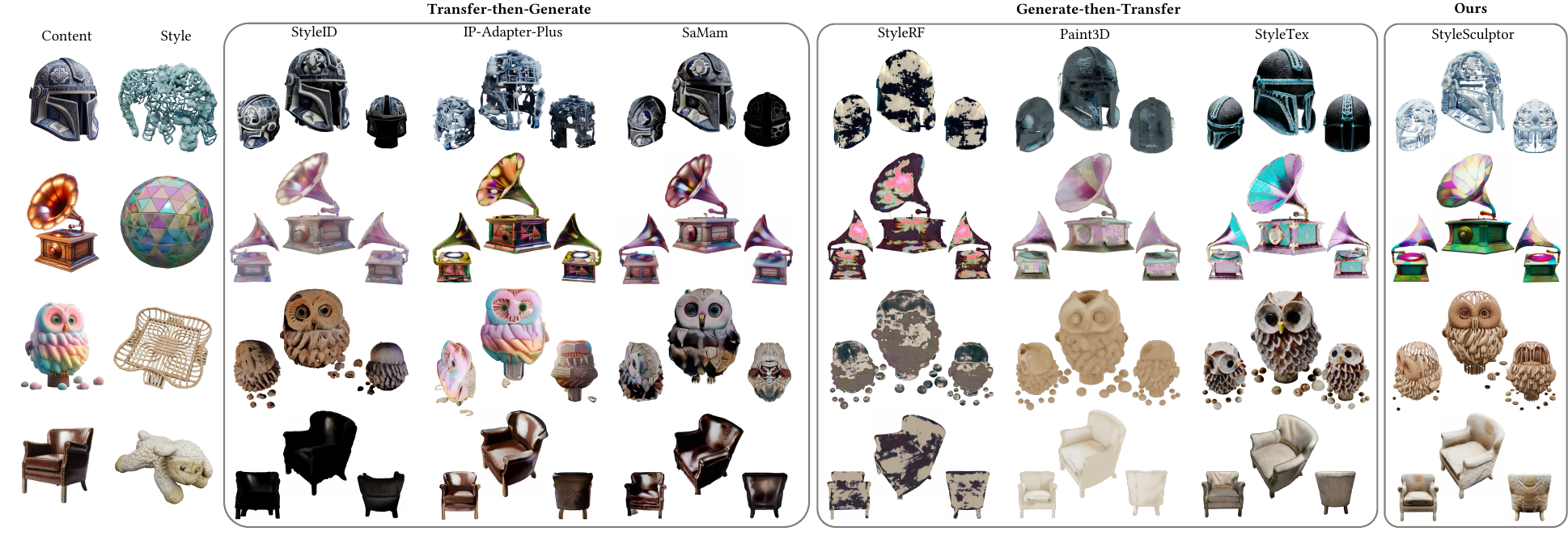}
  \vspace{-5mm}
  \caption{\modify{Qualitative comparison to 2D style transfer method StyleID~\cite{chung2024style}, IP-adapter~\cite{ye2023ip}, SaMam~\cite{liu2025samam} and 3D style transfer method StyleRF~\cite{liu2023stylerf}, Paint3D~\cite{zeng2024paint3d}, and StyleTex~\cite{xie2024styletex}. For all these methods, TRELLIS~\cite{xiang2024structured} is used as the 3D generation backbone. }
  }
  \label{fig:comparison}
  \vspace{-3mm}
\end{figure*}

To the best of our knowledge, \wzw{StyleSculptor is the first approach that enables both texture and geometry} style-guided image-to-3D generation. As none of existing works support our task, we compare StyleSculptor with the two-stage approaches mentioned in Sec.~\ref{intro}: (1) transfer-then-generate approaches and (2) generate-then-transfer approaches.
For \textit{transfer-then-generate}, we adopt StyleID~\cite{chung2024style}, IP-Adapter-Plus~\cite{ye2023ip} \modify{and SaMam~\cite{liu2025samam}} for style transfer on input images, followed by image-to-3D generation using TRELLIS~\cite{xiang2024structured}. For \textit{generate-then-transfer}, we first generate 3D assets with TRELLIS, and then apply 3D style transfer methods, including StyleRF~\cite{liu2023stylerf}, Paint3D~\cite{zeng2024paint3d}, and StyleTex~\cite{xie2024styletex}. Note that StyleTex takes additional content and style text prompts as inputs, which are generated by GPT-4.
To ensure a fair comparison, we use the same single style image as style guidance for all methods, as illustrated in Fig.~\ref{fig:comparison} and~\ref{fig:all_comparison}.

\subsubsection{Qualitative Results.} 
\begin{figure*}[]
  \includegraphics[width=0.99\textwidth, trim=0 460 0 0, clip]{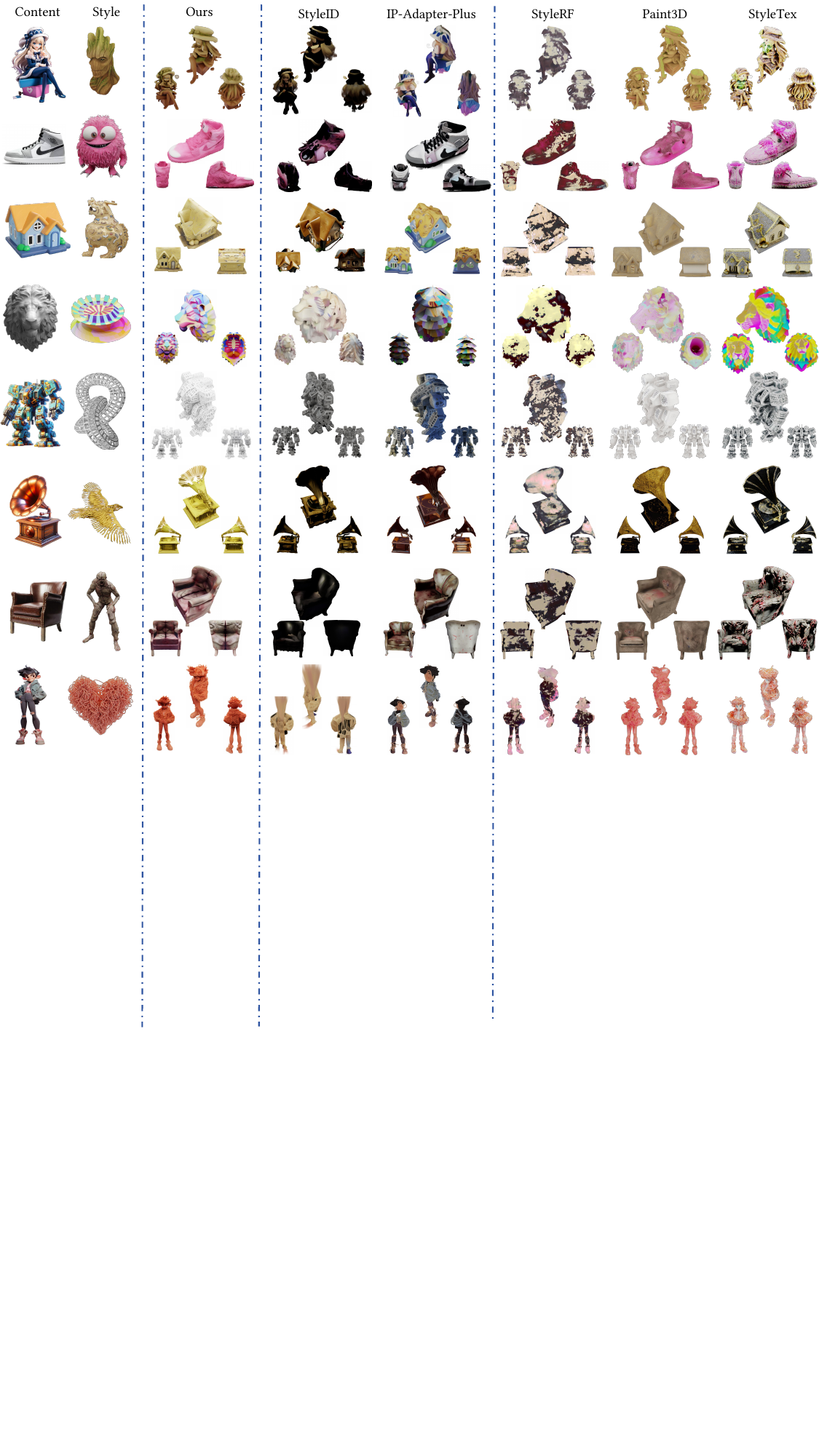}
  \caption{More comparison results of StyleSculptor and sota methods.
  }
  \label{fig:all_comparison}
\end{figure*}

As shown in Fig.~\ref{fig:comparison} and~\ref{fig:all_comparison}, we present extensive comparisons with existing SOTA methods:
(1) For \emph{transfer-then-generate} approaches, \wzw{StyleID} and IP-Adapter-Plus benefits from the strong 2D diffusion \wzw{prior} to capture both texture and geometry style features for 2D style transfer. However, the \wzw{2D transfer} process distorts the original layout of the image, \wzw{resulting in misinterpreted content and semantic loss during} the subsequent 3D generation, such as the back of the helmet and watch, and the flattened appearance of the owl in Fig.~\ref{fig:comparison}.
(2) For \emph{generate-then-transfer} methods, StyleRF mainly targets artistic styles and \wzw{fails to recognize} complex 3D styles, resulting in overall low-quality outputs. Paint3D and StyleTex can generate textures \wzw{consistent} with the style image, but \wzw{struggle with altering} the geometric structure of the 3D asset, \wzw{such as the ignored geometry style} of the helmet. StyleTex relies on text prompts for style control. When the style becomes complex, its performance deteriorates, as shown in the first case of Fig.~\ref{fig:all_comparison}. \wzw{Instead,} our approach enables style-guided 3D generation with geometry-texture dual guidance, while preserving the semantics of the content image, resulting in visually compelling results.

\subsubsection{Quantitative Results.}
We adopt three quantitative metrics on for assessment: 
(1) \textbf{ArtFID}~\cite{wright2022artfid}, which evaluates both content preservation and style transfer quality; 
(2) \textbf{FID} between the output and the style image, measuring the fidelity of style transfer; and 
(3) \textbf{LPIPS} between the output and the content image, indicating the degree of content distortion or style leakage. The results are computed on 60 cases.
\qzf{Please see the supplementary material for the detailed experiment setting.} As shown in Tab.~\ref{tab:quan}, our method achieves superior performance on both ArtFID and FID. While StyleTex benefits from text guidance and shows slightly better content preservation, our method ranks second overall with only a marginal gap in LPIPS, demonstrating strong balance between style transfer and content fidelity. \modify{Even when relying solely on texture or geometry information as guidance, the FID and ArtFID values still increase to a certain extent. This suggests that 3D style generally encompasses both texture and geometry components.}

\begin{table}[t]
\caption{~\textbf{Quantitative evaluation.} All results are tallied across 60 different generation results.}
\vspace{-3mm}
\label{tab:quan}
\centering
\setlength{\tabcolsep}{12pt}
\resizebox{\columnwidth}{!}{%
\begin{tabular}{c|cccl}
\cline{1-4}
\textbf{Method}                        & \textbf{ArtFID}
$\downarrow$ & \textbf{FID}
$\downarrow$ & \textbf{LPIPS}
$\downarrow$   \\ \cline{1-4}
StyleID~\cite{chung2024style}         & 19.94              & 12.18           & 0.5103              \\
IP-Adapter-Plus~\cite{ye2023ip} & \underline{18.39}        & \underline{11.23}      & 0.5073              \\
\modify{SaMam~\cite{liu2025samam}} & \modify{20.44}        & \modify{12.58}      & \modify{0.5055}              \\ \cline{1-4}
StyleRF~\cite{liu2023stylerf}         & 20.06              & 12.43           & 0.4981    \\
Paint3D~\cite{zeng2024paint3d}         & 18.81              & 11.61           & 0.4973            \\ 
StyleTex~\cite{xie2024styletex}        & 20.22              & 12.53           & \underline{0.4961}             \\ \cline{1-4}
  \modify{\textbf{StyleSculptor (Geo-Only)}}          & \modify{18.46}     & \modify{11.46}         & \modify{0.4829}  \\
  \modify{\textbf{StyleSculptor (Tex-Only)}}           & \modify{19.92}     & \modify{12.65}         &\modify{\textbf{0.4617}}  \\
    \textbf{StyleSculptor (Dual-Style)}           & \textbf{17.07}     & \textbf{10.41}         &0.4971  \\ \cline{1-4}
\end{tabular}%
}
\vspace{-2mm}
\end{table}


\subsubsection{User Study.}
\wzw{The metrics used above mainly measure the
input-output similarity, which cannot present the overall performance of different methods. We
thus conduct a user study to estimate real-world user preferences. We invite 30 users publicly to complete a questionnaire for preference
comparisons. Results in~\reftab{userstudy} present that our approach
significantly outperforms existing methods in style-guided 3D generation in terms of human preferences. Detailed settings of user study can be found in the supplementary material.}

\subsection{Ablation Study}

\subsubsection{Settings.}
\begin{figure*}[t]
  \includegraphics[width=\textwidth, trim=0 275 135 0, clip]{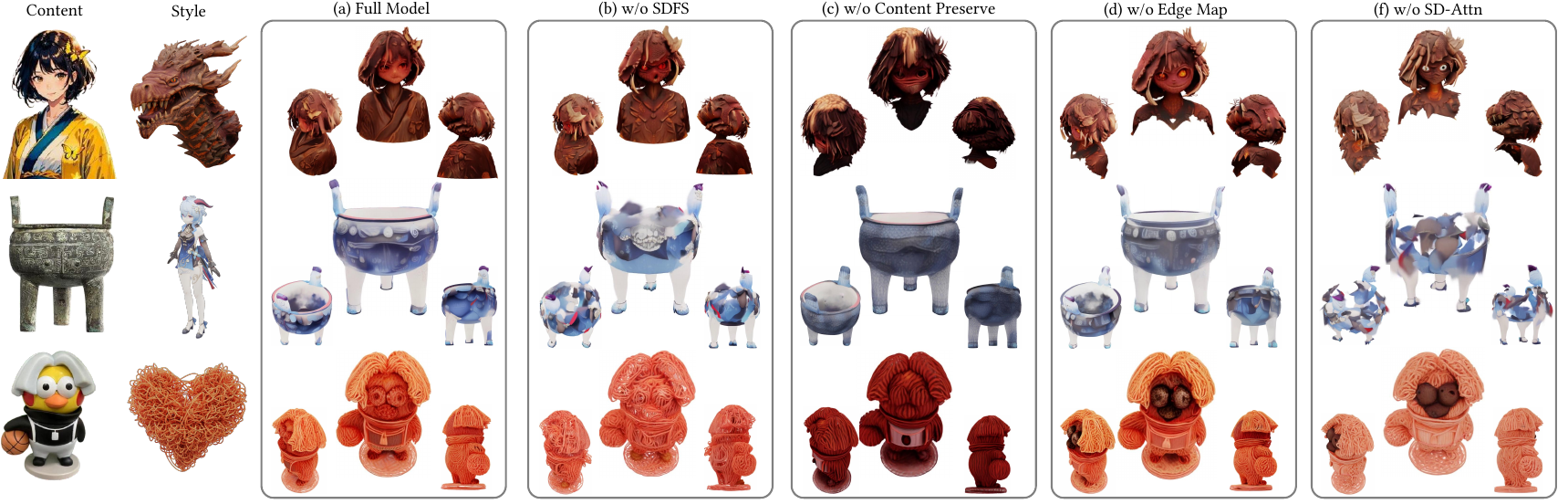}
  \caption{Overall ablation study on StyleSculptor.
  }
  \label{fig:ablation}
\end{figure*}

To evaluate the effectiveness of each component, we conduct an ablation study as shown in Fig.~\ref{fig:ablation}. The variants \wzw{of our approach} are defined as follows: \wzw{
(a) Full Model.
(b) w/o SDFS: replacing the SDFS strategy with a random channel selection strategy.
(c) w/o Content Preserve: removing the content-preserve path and combining self-attention directly with cross-attention.  
(d) w/o Edge Map: computing the style-aware mask using style features instead of edge maps.  
(e) w/o SD-Attn: removing the whole SD-Attn module and simply using content image in stage 1 and style image in stage 2.  }

\begin{table}[]
\caption{~\textbf{User study results.} We present texture preference (TP) and geometry preference (GP) ratio on the same 18 cases.}
\vspace{-3mm}
\label{tab:userstudy}
\setlength{\tabcolsep}{9pt}
\resizebox{0.75\columnwidth}{!}{%
\begin{tabular}{c|ccl}
\cline{1-3}
\textbf{Method}                   & \textbf{TP}
$\uparrow$ & \textbf{GP}
$\uparrow$   \\ \cline{1-3}
StyleID~\cite{chung2024style}                    & 0.19\%                   & 1.48\%                   \\
IP-Adapter-Plus~\cite{ye2023ip}      & 5.56\%            & 6.67\%                   \\ \cline{1-3}
StyleRF~\cite{liu2023stylerf}                   & 0.37\%                  &  1.67\%         \\
Paint3D~\cite{zeng2024paint3d}                    & 12.96\%                  &   11.30\%                \\
StyleTex~\cite{xie2024styletex}                   &  11.30\%                 &  12.41\%                  \\ \cline{1-3}
\textbf{StyleSculptor (Ours)}                   &\textbf{69.63\%}         & \textbf{66.48\%}            \\ \cline{1-3}
\end{tabular}%
}
\vspace{-2mm}
\end{table}

\subsubsection{Results}
\textbf{Effectiveness of SD-Attn.}
As shown in (b) and (f), the absence of the SD-Attn module leads to difficulties in feature fusion when the content and style inputs exhibit large semantic discrepancies (e.g., the vessel and character in the second row), resulting in failed or implausible generation outcomes. This demonstrates \wzw{the effectiveness of the SD-Attn module to integrate semantically distinct features.}
\textbf{Effectiveness of SDFS.} From (a) and (b), the SDFS mechanism effectively preserves the semantic content of the input, such as the eyes and clothing of the chicken toy. 
\textbf{Effectiveness of Content Preserve.} From  (a) and (c), the absence of the content preserve mechanism leads to a loss of semantic content. The features obtained through cross-attention alone do not provide sufficient semantic information, resulting in disrupted feature distributions.  \textbf{Effectiveness of Edge map.} From (a) and (d), when edge map features are used as the filter in SDFS, the channel selection capability is further enhanced by reducing the interference of local content variations. 

\subsection{Insight Validation}

\begin{figure*}
  \includegraphics[width=\textwidth, trim=0 335 220 0, clip]{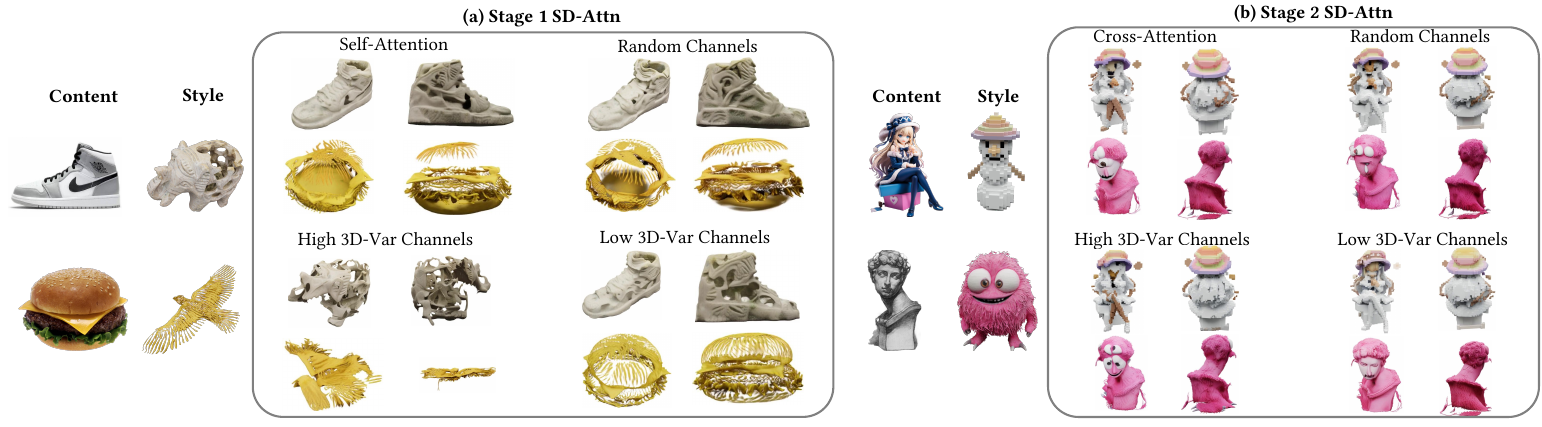}
  \caption{Ablation study on SD-Attn module in StyleSculptor.
  }
  \label{fig:SD-Attn}
\end{figure*}

\modify{The validation experiments of the two insights proposed in Sec.~\ref{sa-att} are shown in Fig.~\ref{fig:SD-Attn} and Tab.~\ref{tab:insight}. For both stages, we evaluate three different channel selection strategies: random selection, selecting channels with the maximum 3D-Var, and with the minimum 3D-Var. When evaluating one stage, the other stage is kept intact. The number of selected channels remains the same across all settings. }

\modify{In \textbf{Stage 1}, random channel selection leads to only minor geometric changes. When using the channel with the highest variance (content-aware channels), the structure collapses significantly and the overall shape becomes more similar to the style asset, which lead to significant increase in the LPIPS value of the generated results. This is because these content-aware channels contain strong spatial information that heavily influences the geometry. In contrast, using channels with lower variance (style-aware channels) achieves a more desirable style-guided effect.}

\modify{In \textbf{Stage 2}, both random channel selection and selecting channels with the highest 3D-Var lead to significant content leakage and our selection mode effectively retains the semantic message of the content image. From the results obtained from two stages of channel selection using 3D-Var, it is evident that channels can be decomposed into style-significant and content-significant, and also that 3D-Var can be a valid mathematical statistic for such selection.
}

\begin{table}[t]
\caption{\modify{~\textbf{Insight Validation.} The ablation results with SD-Attn in different channel selection settings.} }
\vspace{-2mm}
\label{tab:insight}
\centering
\setlength{\tabcolsep}{12pt}
\resizebox{\columnwidth}{!}{%
\begin{tabular}{c|cc|cc}
\cline{1-5}
\textbf{Setting} & \multicolumn{2}{c|}{\textbf{Stage 1}} & \multicolumn{2}{c}{\textbf{Stage 2}} \\
\cline{2-5}
 & \textbf{FID}$\downarrow$ & \textbf{LPIPS}$\downarrow$ & \textbf{FID}$\downarrow$ & \textbf{LPIPS}$\downarrow$ \\
\cline{1-5}
Random channels        & 10.88              & 0.4973           & 11.50  & 0.4980                          \\
High 3D-Var channels & 10.69        & 0.5249      &   11.14          &  0.5032      
\\
Low 3D-Var channels & \textbf{10.41}       & \textbf{0.4971}      & \textbf{10.41}              & \textbf{0.4971}          \\
\cline{1-5}
\end{tabular}%
}
\vspace{-2mm}
\end{table}

\begin{table}[t]
\caption{\modify{~\textbf{Generalization Ability.} The method transfer result of SD-Attn on the Hunyuan3D backbone.} }
\vspace{-2mm}
\label{tab:generalization}
\centering
\setlength{\tabcolsep}{12pt}
\resizebox{\columnwidth}{!}{%
\begin{tabular}{c|cccl}
\cline{1-4}
\textbf{Setting}                        & \textbf{ArtFID}
$\downarrow$ & \textbf{FID}
$\downarrow$ & \textbf{LPIPS}
$\downarrow$   \\ \cline{1-4}
Backbone (S1-Content, S2-Style)        & 22.95             & 14.14           & 0.5159              \\
+ Cross-3D Attention & 21.74        & 13.50    & 0.5060              \\
+ SD-Attn (High 3D-Var) & 20.93        & 12.96      & 0.5092              \\ 
+ SD-Attn (Low 3D-Var) & \textbf{20.01}         & \textbf{12.38}      & \textbf{0.4979}              \\ \cline{1-4}
\end{tabular}%
}
\vspace{-2mm}
\end{table}

\subsection{Generalization Ability}



\begin{figure}[h]

  \includegraphics[width=0.49\textwidth, trim=10 430 620 5, clip]{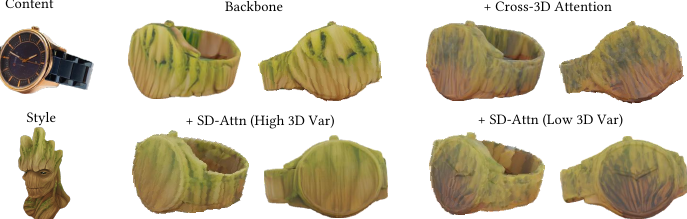}
  \vspace{-6mm}
  \caption{\modify{The generalization experiment on Hunyuan3D backbone.}
  }
  \label{fig:hunyuan}
   \vspace{-2mm}
\end{figure}

\modify{To demonstrate the generalization ability of our method, we replace the original backbone TRELLIS with Hunyuan3D-2.1~\cite{hunyuan3d2025hunyuan3d} and modify its stage-2 architecture by substituting the original self-attention computation in the down, middle, and up transformer layers with our proposed SD-Attn module. The qualitative and quantitative results are shown in Fig.~\ref{fig:hunyuan} and Tab.~\ref{tab:generalization}.}
\modify{These results show (1) consistent improvement in style-guided generation through fine-grained 3D feature fusion of Cross-3D Attention (Tab.~\ref{tab:generalization}, 1st and 2nd rows); and (2) effectiveness of variance selection in separating style- and content-significant features (Tab.~\ref{tab:generalization}, 3rd and 4th rows). These results indicate that our method is not tied to a specific backbone, suggesting its compatibility with various 3D generative models.}

\section{Application}
\begin{figure*}
  \includegraphics[width=\textwidth, trim=0 375 75 0, clip, keepaspectratio]{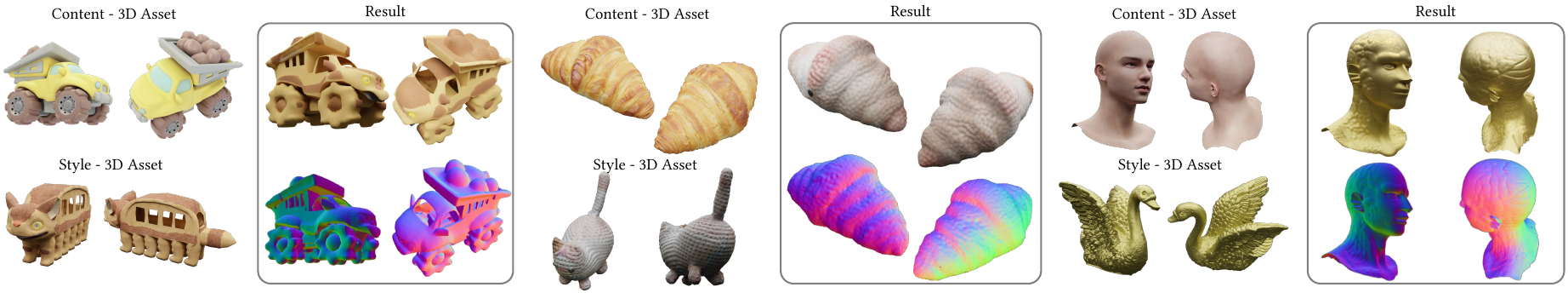}
        \caption{\modify{Qualitative results of Geometry-Enhanced 3D Style Transfer.}}
        \label{fig:3d_style}
\end{figure*}

\textbf{Geometry-Enhanced 3D Style Transfer.} In addition to style-guided \wzw{image-to-3D} generation, \wzw{StyleSculptor} can also \wzw{be extended to conduct 3D style transfer} by taking \wzw{the rendered images of an 3D asset as content input and another 3D asset as style input}. \wzw{We show the} results in Fig.~\ref{fig:3d_style}. Unlike previous 3D style transfer methods that only alter the textures of an existing 3D model, StyleSculptor supports fine-grained geometric control while keeping the semantic structure of the content 3D asset unchanged.

\section{Conclusion}
This paper introduces \textit{StyleSculptor}, the first \wzw{style-guided 3D generation approach} that supports \wzw{texture-geometry dual guidance}. StyleSculptor incorporates a novel SD-Attn module, which integrates cross-attention into the 3D latent feature space to effectively inject style \wzw{controls} during the generation process. To address the content leakage issue, the SD-Attn module leverages the variance of 3D feature patches to disentangle style-significant channels from the full feature space, enabling selective feature injection within the attention framework. Finally, we propose a Style Guided Control (SGC) mechanism that allows users to flexibly control the intensity and \wzw{granularity} of style guidance, achieving user-controllable style-guided 3D generation.

\begin{figure}[h]
\vspace{-3mm}

  \includegraphics[width=0.48\textwidth, trim=0 365 475 0, clip]{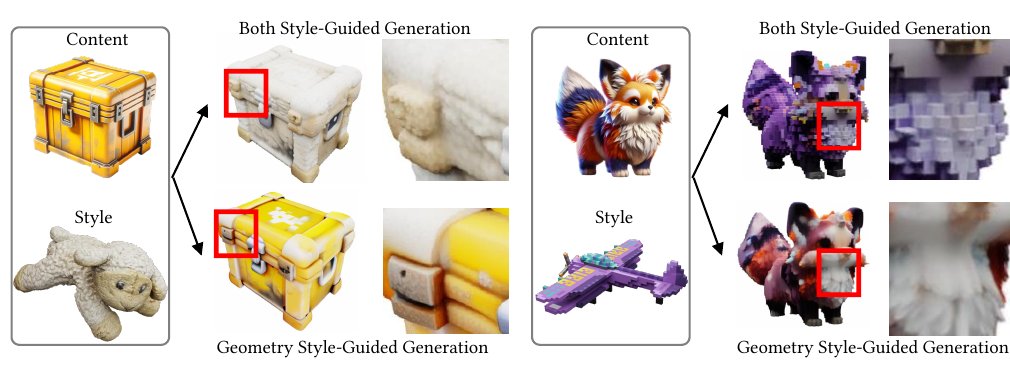}
  \vspace{-6mm}
  \caption{\modify{Failure cases.}
  }
  \label{fig:failure}
   \vspace{-2mm}
\end{figure}

\textbf{Limitations.}
As shown in Fig.~\ref{fig:failure}, StyleSculptor \wzw{fails to perform geometry-only style-guided generation} when the target style is subtle or localized, such as furry surfaces or pixel-art designs. This limitation primarily stems from the strong coupling between local geometric details and texture information in the TRELLIS latent feature space, making it challenging for the feature disentanglement and fusion operations to handle such complex cases effectively. \modify{It may be mitigated by adopting a stronger baseline or incorporating some geometric priors (e.g., curvature constraints), which will be explored in future work.} Nevertheless, StyleSculptor is still capable of achieving dual-style guided generation under these conditions.

\bibliographystyle{ACM-Reference-Format}
\bibliography{arxiv}


\clearpage
\clearpage
\section{Single-Style Guided Method}
For texture-only style guidance, a small $ K $ suffices to preserve content semantics while guiding only the style of textural appearance. 
For geometry-only style guidance, we first perform generation with a large $ K $ to capture both texture and geometry styles. Then, we render the generated asset and re-run StyleSculptor with the rendering images as new content image, and use the original $I_c$ as the new style image with a smaller $ K $. In this re-run procedure, the SD-Attn layer in the first stage is disabled  — only self-attention on the content input is performed, ensuring no significant geometry change. This process preserves the geometric style information and recolors the asset using the original content image.

\section{Implemention Details}
 \subsection{Backbone.} StyleSculptor is built upon the TRELLIS~\cite{xiang2024structured} backbone. For all experiments, we use the publicly available \textit{TRELLIS-image-large} model and its pre-trained weights provided in the official codebase. We replace the self-attention layers in every Transformer block of the first two stages with our proposed SD-Attn modules, while keeping the rest of the network architecture unchanged.
All model weights are frozen during inference, and no fine-tuning or parameter updates are required. Style-guided 3D generation is achieved through a two-stage feed-forward inference process. In Stage 1, the classifier-free guidance scale (CFG) is set to 6.5 with 100 sampling steps; in Stage 2, the CFG is set to 3.5, also with 100 steps. The number of style-aware channels, $ K $, is set to 80 in Stage 1 and 800 in Stage 2.
For visualization, all colorized 3D results in the paper are rendered using Gaussian Splatting based on the outputs of TRELLIS. Mesh visualizations are generated using normal maps extracted from the predicted 3D mesh.

\subsection{Edge Map.} 
In this work, we employ the PidiNet~\cite{su2021pixel} network to extract edge maps, which are fed into the model at the same resolution as the original style images. When pre-processing the style images using the TRELLIS pipeline default functions (e.g., background removal, resizing), we record the applied transformation parameters and apply the same operations to the edge maps, ensuring that both modalities retain consistent semantic information.

\section{Experiment Setting}
\subsection{Quantitative Experiment.} In our quantitative experiment, we select 15 content images and randomly pair each with 4 out of 20 style assets that exhibit clear stylistic differences, resulting in a total of 60 generated samples. For each result, we use the main view image under the same rendering configuration to compute the evaluation metrics.

\subsection{User Study.}
Besides the quantitative results, we conduct a user study to further evaluate the dual-style guided generation ability of our method. For this study, we collect 18 test cases and 36 questions with identical input across all methods, shuffle the outputs, and invite 30 public participants to vote on which result has the best geometric and texture style attribute in each case. 
The preference results are summarized in Tab. 2. As shown, our method achieves a significantly higher preference ratio than competing approaches, demonstrating its superiority in generating 3D stylized asset.

\section{Assets Attribution.}
In this paper, we use 3D models sourced from the ObjaverseXL dataset~\cite{deitke2023objaverse} and Sketchfab\footnote{https://sketchfab.com.} under the Creative Commons Attribution 4.0 International (CC BY 4.0) license. The geometry and texture information of these model are not modified and just be used as the style guidance in our paper. All 3D assets obtained from Sketchfab have been confirmed that have no \textit{NoAI} item.

Each model used from Sketchfab is attributed as follows:
\begin{itemize}
\item[$\bullet$] “PIXEL PLANE” by LordCinn. 
\item[$\bullet$] "a colorful and abstract creature with tentacles" by klrxyz.
\item[$\bullet$] "Silver Tiling" by Aiekick.
\item[$\bullet$] "Monster Skeleton 2" by Aiekick.
\item[$\bullet$] "Vessel decorated with garlands" by Virtual Museums of Małopolska.
\item[$\bullet$] "The Thunderman dance" by MAXDESIGN-3D.
\item[$\bullet$] "Plexus" by Selfburning.
\item[$\bullet$] "Zinnias in a decorated tile vase" by Advanced Visualization Lab - Indiana University.
\item[$\bullet$] "Plexus" by Selfburning.
\item[$\bullet$] "Contemporary Abstract Sculpture" by curpeja.
\item[$\bullet$] "owlbear-3d-model" by AidanYT55Twt.
\item[$\bullet$] "Dead space armor" by photon (that one larry).
\item[$\bullet$] "14 11 2022 15 12 V0" by openroomxyz.
\item[$\bullet$] "Harlequin\_ Orb" by Egypt VR.
\item[$\bullet$] "Dragon Head" by Moon Cube Studios.
\item[$\bullet$] "Ganyu fbx with textures" by \_INSTICT\_.
\item[$\bullet$] "Heart made of strings" by Blenderkurt.
\item[$\bullet$] "Golden Eagle" by Argonaut.
\item[$\bullet$] "Abstract 2- Torus Knot" by Mike Rowley.
\item[$\bullet$] "Ahri KDA - LoL - 3d Printable Model" by Printed Obsession.
\item[$\bullet$] "Modern Art 008" by ParuthidotExE.
\item[$\bullet$] "CatBus Mi Vecino Totoro" by Acalli Twiss.
\item[$\bullet$] "Swan" by Diana Liu.
\item[$\bullet$] "Golden Croissant 3D Model" by Ati.
\item[$\bullet$] "Amigurumi cat" by profjuan.
\item[$\bullet$] "Mythical Beast Censer, c. 1736-1795 CE" by Minneapolis Institute of Art.

\end{itemize}

All content images used in this paper are selected from the StyleBench~\cite{gao2024styleshot} which exhibit explicit 3D structure, as well as images provided in the official TRELLIS~\cite{xiang2024structured} code repository. All of these images are open source for academic research purposes.

\end{document}